\documentclass{article}
\usepackage{spconf,amsmath,graphicx}
\usepackage{booktabs}
\usepackage{hyperref}
\usepackage{marvosym}
\usepackage{graphicx}
\usepackage{amsmath}
\usepackage{amssymb}
\usepackage{booktabs}
\usepackage{algorithm}
\usepackage{algorithmic}
\usepackage{graphicx}
\usepackage{float}
\usepackage{subfig}

\usepackage{amsmath,amssymb}
\usepackage[capitalize]{cleveref}
\crefname{section}{Sec.}{Secs.}
\Crefname{section}{Section}{Sections}
\Crefname{table}{Table}{Tables}
\crefname{table}{Tab.}{Tabs.}

\title{TAOTF: A Two-stage Approximately Orthogonal Training Framework in Deep Neural Networks}
\name{Taoyong Cui\textsuperscript{1}, Jianze Li\textsuperscript{2}, Yuhan Dong\textsuperscript{1}, Li Liu$^{\href{liuli@cuhk.edu.cn}{\textrm{\Letter}}}$\textsuperscript{2}}

\address{\textsuperscript{1}Shenzhen International Graduate School, Tsinghua University, Shenzhen, China \\\textsuperscript{2}Shenzhen Research Institute of Big Data, the Chinese University of Hong Kong, Shenzhen, China}
\begin{document}
%
\maketitle
\begin{abstract}
The orthogonality constraints, including the hard and soft ones, have been used to normalize the weight matrices of Deep Neural Network (DNN) models, especially the Convolutional Neural Network (CNN) and Vision Transformer (ViT), to reduce model parameter redundancy and improve training stability. However, the robustness to noisy data of these models with constraints is not always satisfactory.
In this work, we propose a novel two-stage approximately orthogonal training framework (TAOTF) to find a trade-off between the orthogonal solution space and the main task solution space to solve this problem in noisy data scenarios. In the first stage, we propose a novel algorithm called polar decomposition-based orthogonal initialization (PDOI) to find a good initialization for the orthogonal optimization. In the second stage, unlike other existing methods, we apply soft orthogonal constraints for all layers of DNN model. We evaluate the proposed model-agnostic framework both on the natural image and medical image datasets, which show that our method achieves stable and superior performances to existing methods.
\end{abstract}

\section{Introduction}
\label{sec:intro}
In the past decades, \emph{Deep Neural Network} (DNN) models, especially the \emph{Convolutional Neural Network} (CNN) and \emph{Vision Transformer} (ViT), have developed rapidly in the computer vision field. 
Although these models can automatically learn the hidden deep features from images, there still exist several problems with them. For example, the parameterization or model capacity utilization is insufficient, gradient explosion or disappearance, and there exists significant redundancy among different feature channels \cite{wang2020orthogonal}.

  
In view of this, orthogonality constraints, including the hard and soft ones, were recently used in the field of deep learning to improve model performance. When the filters are learned to be as orthogonal as possible, they become irrelevant and reduce the redundancy of learning features \cite{wang2020orthogonal}. 
Then the model capacity is made full use of, and the ability of feature expression is improved as well. 
For example, a hard orthogonality constraint was imposed in CNN \cite{2016arXiv161105927H}, and retraction-based Riemannian optimization algorithms were used to solve it. 
A soft orthogonality constraint was imposed in CNN \cite{wang2020orthogonal} with an orthogonal penalty loss. 

However, in these works, inappropriate orthogonal constraints are often imposed, ignoring a more important advantage of orthogonal constraints: robustness to noise samples (\emph{e.g.}, noise, blur, exposure and so on). Applying appropriate orthogonal constraints can make each layer of the model closer to a 1-Lipschitz function. Given a small perturbation to input $\Delta x$, the change of output $\Delta y$ is bounded to be low. Therefore, the model enjoys robustness under noisy data.
for example, if we only use the hard orthogonality constraint, one issue is that the solution set of the primary optimization objective does not necessarily intersect with the hard orthogonality constraints \cite{2022arXiv220112133F}. In other words, if the weight matrices of these models are too close to orthogonal matrices, the performance may be worse. Meanwhile, the computing cost of the hard constraints is always expensive.
On the other hand, if we only use the soft orthogonality constraint, it is ineffective to make the weight matrix orthogonal enough, and thus reduces layers' 1-Lipschitz property. More detailed explanations about these issues will be presented in \cref{subsec:why_sec3}.


In this work, to solve the above issues, we propose a \emph{two-stage approximately orthogonal training framework} (TAOTF) to find the trade-off between the Stiefel manifold and the main task solution set. 
More specifically, in the first stage, we propose a novel algorithm called \emph{polar decomposition-based orthogonal initialization} (PDOI) to find a starting point in the Stiefel manifold. This process is somewhat similar to the hard orthogonal constraints but with a much smaller computational cost. Then, in the second stage, we implement soft orthogonal regulation with an orthogonal penalty loss (\emph{e.g.}, \emph{spectral restricted isometry property} (SRIP) \cite{bansal2018can}) on all layers of DNN, and use a common European optimizer (\emph{e.g.}, Adam) to find an optimal point. We train the CNN and ViT models using the proposed TAOTF framework and then evaluate these TAOTF-based models on both natural and medical images.

To evaluate the robustness of our framework compared to other methods, we simulate possible data challenges with datasets and conduct comparative experiments, which demonstrate the superiority of our framework. 
Except for these two models, this novel framework can be also used together with other DNN models, \emph{e.g.}, the \emph{Recurrent Neural Network} (RNN), to further improve the robustness performances in real scene datasets.

In summary, three contributions can be summarized as follows: 
\begin{itemize}
\item[(1)] We propose a novel model-agnostic framework called TAOTF by combining the advantages of soft and hard orthogonality constraints to improve the robustness performances in DNNs, especially the CNN and ViTs. 

\item[(2)] In the first stage of TAOTF, to find a good starting point for orthogonal optimization, we propose a novel algorithm called PDOI to search near the initial point and update parameter matrices.
\item[(3)] We also conduct extensive experiments in many image datasets, including natural and medical ones. The experimental results show that TAOTF-based models have better robustness performances to noisy perturbations than existing methods.
\end{itemize}
\begin{figure}[tbp]
	\begin{center}
		\includegraphics[width=200pt]{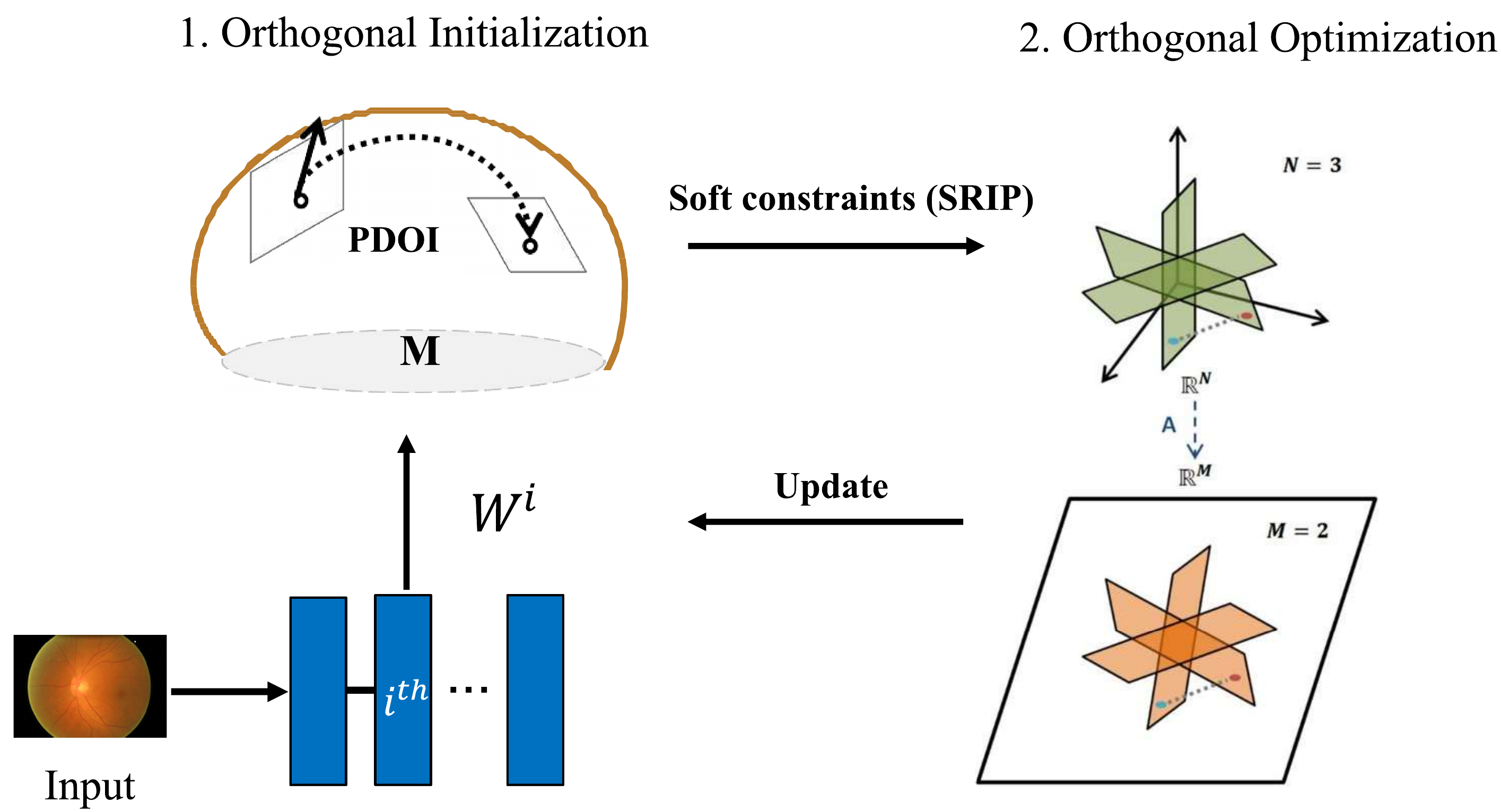}
	\end{center}
	\caption{Our proposed two-stage orthogonal training framework (TAOTF). In the first stage, we use PDOI to find a good starting point on Stiefel Manifold by iterating. In the second stage, we impose soft orthogonal constraints on all layers of DNN to find a trade-off.}
	\label{fig:TAOTF}
\end{figure}
\section{Related Works}
\label{sec:related}

In this section, we mainly review some related works in the literature about the applications of hard and soft orthogonality constraints in the DNN models, especially the CNN and Transformers.

\subsection{Hard orthogonality constraints}
To our knowledge, the first work using hard orthogonality constraints in CNN was \cite{2016arXiv161105927H}, where the Stiefel layer was introduced, and Riemannian optimization techniques on matrix manifolds were used in AlexNet and VGG. 
Then, a new backpropagation with a variant of \emph{stochastic gradient descent} (SGD) on Stiefel manifolds \cite{huang2017riemannian} was exploited to update the structured connection weights. 
In \cite{2017arXiv170906079H}, the authors generalized such square orthogonal matrices to rectangular ones, and formulated this problem in \emph{Feed-forward Neural Networks} (FNNs) as an optimization problem over multiple dependent Stiefel manifolds. 
Recently, in \cite{2019arXiv190108428L}, an alternative approach was proposed based on a parameterization stemming from Lie group theory, and the constrained optimization problem was transformed into an unconstrained one over a Euclidean space.

\subsection{Soft orthogonality constraints}
The soft orthogonality constraints were mainly solved using a penalty loss function calculating the discrepancy between the identity matrix $\boldsymbol{I}$ and the product of weight matrix $\boldsymbol{W}$ and its transpose $\boldsymbol{W^{T}}$, \emph{e.g.}, $\|\boldsymbol{W}\boldsymbol{W^{T}}-\boldsymbol{I}\|$.
To our knowledge, the first work using this soft orthogonality constraint in training DNNs was \cite{2012arXiv1211.5063P}, where it was used in \emph{Recurrent Neural Networks} (RNNs) to help avoid
gradient vanishing/ explosion, 
and the first work using the soft orthogonality constraint in CNNs was \cite{2017arXiv170301827X}, which helped to stabilize the layer-wise distribution of activations. In \cite{bansal2018can}, to enforce the orthogonality regularizations, the authors used a novel regularization form for orthogonality in CNNs, named \emph{Spectral Restricted Isometry Property} (SRIP).
In \cite{wang2020orthogonal}, a new \emph{orthogonality based CNN} (OCNN) was proposed, and good results on multiple natural datasets were achieved approving their robustness under attack. In \cite{GEHLOT2021102099}, class vectors were applied to improve the ability of the model to resist the label noise of datasets for cancer diagnosis. 

In transformers, orthogonal weights can also improve numerical stability during training and upper-bound the Lipschitz constant of linear transformations. In \cite{zhang2021orthogonality}, they first applied the basic orthogonality constraint on transformers and achieved good results in several NLP tasks such as neural machine translation and sequence-to-sequence dialogue generation. 
In \cite{2022arXiv220112133F}, they developed an orthogonal Vision Transformer (O-ViT), which also used methods like \cite{2019arXiv190108428L} to impose orthogonality constraints on self-attention layers.

\section{Methods}
In this section, we will detailedly introduce the new training framework TAOTF (\cref{fig:TAOTF}). 
In \cref{subsec:why_sec3}, we will explain the reason why we propose a new two-stage orthogonal training framework.
In \cref{subsec:stage1_maniopt}, the proposed algorithm PDOI to find a good optimization starting point in the first stage will be introduced. In \cref{subsec:stage2_maniregu}, we will introduce the soft constraints we use in the second stage. 

\subsection{Why we need a two-stage framework?}\label{subsec:why_sec3}

In a DNN model, it is well known that, if $\boldsymbol{X}$ is the weight matrix of the $i$-th layer, then the training of $\boldsymbol{X}$ is to solve the following optimization problem:
\begin{equation}\label{eq:opt_Rnp_f}
\min_{\boldsymbol{X}\in\mathbb{R}^{n\times p}} g(\boldsymbol{X})
\end{equation}
where $g$ is the loss function. 
Let 
$$\textbf{St}(p,n) \stackrel{\sf def}{=}\{\boldsymbol{X}\in\mathbb{R}^{n\times p}: \boldsymbol{X^{T}}\boldsymbol{X}=\boldsymbol{I}_p\}$$
be the \emph{Stiefel manifold}, 
where $1\leq p\leq n$, and $\boldsymbol{I}_p$ denotes the identity matrix of size $p$. 
As explained in \cref{sec:intro}, to reduce the redundancy of learning features, we would like to impose orthogonality on the weight matrix $\boldsymbol{X}$.

One approach is to use a \emph{hard orthogonality constraint}, and then problem \eqref{eq:opt_Rnp_f} becomes a Riemannian optimization problem \cite{absil2009optimization} on the Stiefel manifold \cite{2020arXiv200201113L},
\emph{i.e.}, 
\begin{equation}\label{eq:opt_steif_f}
\min_{\boldsymbol{X}\in\textbf{St}(p,n)} g(\boldsymbol{X}).
\end{equation}
In the iterations of the algorithm to solve problem \cite{absil2009optimization}, the weight matrix $\boldsymbol{X}$ will always stay in $\textbf{St}(p,n)$, and thus can be kept columnly orthogonal. 
The other approach is to use a \emph{soft orthogonality constraint}, and then problem \eqref{eq:opt_Rnp_f} becomes
\begin{equation}\label{eq:opt_Rnp_f_soft}
\min_{\boldsymbol{X}\in\mathbb{R}^{n\times p}} g(\boldsymbol{X})+\lambda r(\boldsymbol{X}),
\end{equation}
where $\lambda>0$ and $r(\boldsymbol{X})$ is a regularization term to enforce the orthogonality of $\boldsymbol{X}$.

\begin{figure}[tbp]
\centering
\subfloat{\includegraphics[width=0.4\textwidth]{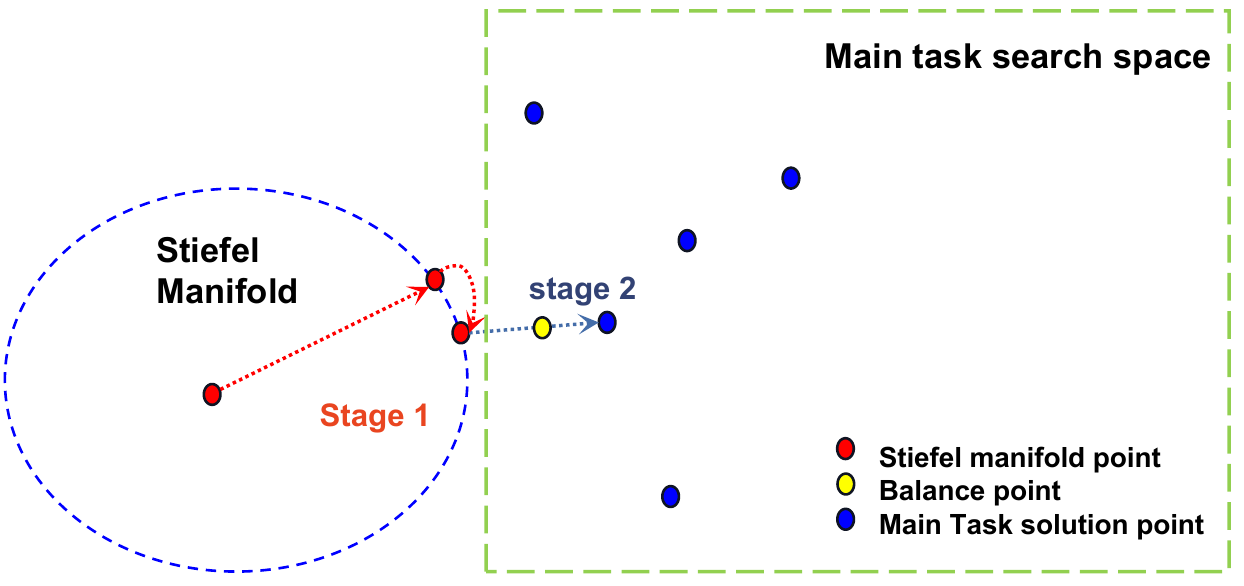}}
\caption{Stiefel manifold and the solution set. The solution set of the primary optimization objective does not necessarily intersect with the Stiefel manifold, our method can find a balance point between them.} 
\label{Stiefel manifold}
\end{figure}

As introduced in \cref{sec:related}, the above hard and soft orthogonality constraints were both used to improve the robustness performances of DNN or ViT models. 
However, as explained in \cref{Stiefel manifold}, the solution matrices of problem \eqref{eq:opt_Rnp_f} maybe not be in $\textbf{St}(p,n)$. Therefore, if we only use the hard orthogonality constraint to solve problem \eqref{eq:opt_steif_f}, the solution may be too restrictive. In other words, the solution set of the primary optimization objective does not necessarily intersect with the hard orthogonality constraints \cite{2022arXiv220112133F}, and thus the performance of the trained DNN models may degrade. 
On the other hand, if we only use the soft orthogonality constraint to solve problem \eqref{eq:opt_Rnp_f_soft}, the solution may be far away from $\textbf{St}(p,n)$, and thus the DNN models may still suffer from the parameter redundancy, and it is not robust enough as well. 

Enforcing proper orthogonality constraints can generate a more uniform spectrum \cite{wang2020orthogonal}, which makes network layers closer to a 1-Lipschitz function like 
\begin{equation}\label{Lipschitz}
\left\|f\left(x_{1}\right)-f\left(x_{2}\right)\right\| \leq \left\|x_{1}-x_{2}\right\|.
\end{equation} Given a slight perturbation to the input $\Delta x$, the variation of the output $\Delta y$ is bounded low, producing a robust and less sensitive representation of data perturbations.

Therefore, in this work, we propose a novel TAOTF framework, which includes two stages at each iteration to find the trade-off between the search space of the main task and the orthogonality constraint. From a perspective of optimization theory, it can be understood that we first solve problem \eqref{eq:opt_steif_f} to calculate a good starting point, and then solve problem \eqref{eq:opt_Rnp_f_soft}. 
It will be seen in \cref{sec:experim_data} that, although the TAOTF framework includes two stages, it still has competitive convergence speed. 
One reason is the use of a projection-based PDOI algorithm in the first stage, which does not need to calculate the retraction map. 
The other reason is that we control the iteration numbers at the first stage.

\subsection{First Stage: Orthogonal Initialization }\label{subsec:stage1_maniopt}

To solve the problem \eqref{eq:opt_steif_f}, the retraction-based optimization algorithm \cite{absil2009optimization} was proposed in recent years. However, as the retraction-based algorithms are generally expensive, in this work, inspired by low-rank orthogonal approximation of tensors \cite{chen2009tensor}, we propose a novel algorithm PDOI in \cref{alg:algorithm}, to find a local optimum as the starting point for orthogonal optimization of the second stage. Although the global convergence of the $X_k$ is not determined, under this mild condition, the input point can be converged (locally) to an extreme point by the PDOI algorithm nearby, which can be used as the starting point for the next stage. The initial point found in this way, on the one hand, satisfies the constraint of the Stifel manifold, and on the other hand, finds a point closer to the main task search space on the Stifel manifold with low computing costs.

\begin{algorithm}[!htbp]
\caption{PDOI algorithm}
\label{alg:algorithm}
\textbf{Input}: a starting point $\boldsymbol{X}_0$, a positive constant $\gamma>0$.\\
\textbf{Output}: $\boldsymbol{X}_k$, $k \geq 1$.
\begin{algorithmic}[1] 
\FOR{k=1,2,..., until a stopping criterion is satisfied}
\STATE Compute $\nabla g\left(\boldsymbol{X}_{k-1}\right)$.
\STATE Compute the SVD decomposition ($\boldsymbol{U} \boldsymbol{\Sigma} \boldsymbol{V^T}$) of 
\begin{equation}\label{eq:pd_item_grad}
\nabla g\left(\boldsymbol{X}_{k-1}\right)+\gamma \boldsymbol{X}_{k-1}.
\end{equation}
\STATE Update $\boldsymbol{X}_k$ to be the product of two orthogonal matrices $\boldsymbol{U} \boldsymbol{V^T}$.
\ENDFOR
\end{algorithmic}
\end{algorithm}

The PDOI algorithm employs an alternating procedure (iterating through $X_0$, $X_1$,...,$X_k$,...,$X_N$), where in each step all but one ($X_n$) parameters are fixed. In general, algorithms of this type, including alternating least squares, are not guaranteed to global convergence, but the iterations can search for points closer to the main task on the Stiefel manifold. Moreover, of the generated parameter sequence, every converging subsequence converges to a stationary point of the objective function, which can be a good starting point from the perspective of optimization theory.

As proved in \cite{hu2020provable}, the initial weights
from the orthogonal group not only speeds up convergence relative to the standard Gaussian initialization with iid weights but close to isometry during training to enable efficient convergence and the 1-Lipschitz property. The algorithm PDOI not only guarantees that the starting point must be on the Stiefel manifold (initial weights in the orthogonal group), but, through several iterations, can find the better starting point near the input point suitable for both orthogonal constraints and main tasks. Extensive experiments have been conducted to prove this view.

\subsection{Second Stage: Orthogonal Optimization}\label{subsec:stage2_maniregu}
Recall that the \emph{Restricted Isometry Property} (RIP) condition of a matrix $\boldsymbol{W}\in\mathbb{R}^{n\times n}$ means that, for all vectors $\boldsymbol{x}\in\mathbb{R}^{n}$ that are $k$-sparse, there exists a small $\delta_{\boldsymbol{W}}\in(0, 1)$ such that
\begin{equation}\label{eq:rip}
\left(1-\delta_{\boldsymbol{W}}\right)\|\boldsymbol{x}\|_{2}^{2} \leq \|\boldsymbol{W} \boldsymbol{x}\|_{2}^{2}  \leq\left(1+\delta_{\boldsymbol{W}}\right)\|\boldsymbol{x}\|_{2}^{2}.
\end{equation}
The positive constant $\delta_{\boldsymbol{W}}$ in equation \eqref{eq:rip} is called the \emph{constrained isometric constant}.
If $\delta_{W}$ is very small, it can be interpreted as those $ k$ columns are approximately orthogonal. 
If the equation \eqref{eq:rip} is satisfied with $\delta_{\boldsymbol{W}}=0$ for all $k$-sparse vectors $\boldsymbol{x}\in\mathbb{R}^{n}$, then the matrix $\boldsymbol{W}\in\mathbb{R}^{n\times n}$ satisfies the \emph{isometric characteristics of k-order constraints}. 
The RIP can be used to measure the similarity between the subset composed of $k$ columns in a matrix and an orthogonal matrix. 

The extreme case with $k=n$ was also considered in \cite{bansal2018can}, where the RIP condition will force the whole matrix $\boldsymbol{W}$ to be very close to an orthogonal one, \emph{i.e.}, 
\begin{equation}\label{eq:srip_cond}
\left|\frac{\|\boldsymbol{W} \boldsymbol{x}\|^{2}}{\|\boldsymbol{x}\|^{2}}-1\right| \leq \delta_{\boldsymbol{W}}, \forall \boldsymbol{x} \in \mathbb{R}^{n}.
\end{equation}
In this case, the RIP condition \eqref{eq:srip_cond} is termed as the \emph{Spectral Restricted Isometry Property} (SRIP) regularization.

Existing methods \cite{wang2020orthogonal,zhang2021orthogonality,bansal2018can,2022arXiv220112133F,2016arXiv161105927H,2021arXiv210412753G} always only impose orthogonality constraints on the deep convolutional layers or the self-attention layer of Transformers. However, if we only impose orthogonality constraints on only some layers of the network, the different levels of orthogonality will destroy the 1-Lipschitz property of the global model and damage model robustness against small perturbations. And we proved this view through extensive experiments in \cref{sec:experim_data}. Therefore, we impose soft constraints like SRIP regulation \eqref{eq:srip_cond} on all layers of the model to solve the orthogonal optimization in the second stage.

This process is to separate the starting point found in the first stage from the Stiefel manifold, and further explore the main task solution space but is still limited by the Stiefel manifold. Through such a process, the trade-off of the main task and the orthogonal constraints can be well found, and the final loss function in \eqref{eq:opt_Rnp_f_soft} is
\begin{equation}
\min_{\boldsymbol{X}\in\mathbb{R}^{n\times p}}g(\boldsymbol{X})=g_{M}(\boldsymbol{X})+\lambda g_{SRIP}(\boldsymbol{X}).
\end{equation}
\begin{center}
\begin{table*}[t!]
  \centering

    \resizebox{\linewidth}{!}{\begin{tabular}{c|c|ccc|ccc|ccc|ccc}
        \hline
    \multicolumn{14}{ c }{\textbf{Experiment on Noisy APTOS 2019}}\\
    \hline

              & Clean      &       &    Noise   &       &     & Blur       &       &       &  Weather     &       &       & Digital \\
         \cline{0-13}
          \multicolumn{1}{c|}{Methods}& \multicolumn{1}{c|}{Clean}& \multicolumn{1}{c}{Gaussian.} & \multicolumn{1}{c}{ISO.} & \multicolumn{1}{c|}{Multiplicative.} & \multicolumn{1}{c}{Gauss.} & \multicolumn{1}{c}{Median}
          & \multicolumn{1}{c|}{Motion} & \multicolumn{1}{c}{Optical}  & \multicolumn{1}{c}{Rotate} &\multicolumn{1}{c|}{RGB} & \multicolumn{1}{c}{Bright} & \multicolumn{1}{c}{Frog}& \multicolumn{1}{c}{Saturation} \\

    \multicolumn{1}{c|}{ResNet} &$91.17\%$& $87.32\%$&$89.13\%$  &   $87.05\%$    & $75.60\%$& $63.80\%$ &    $78.14\%$   &  $79.86\%$          &  $64.43\%$ &$90.22\%$    &  $81.60\%$   &$63.53\%$  & $83.06\%$  \\
    ResNet+SRIP \cite{bansal2018can} &   $91.17\%$&  $89.58\%$     &   $89.16\%$    &  $87.95\%$     & $76.12\%$      &  $64.70\%$     &  $77.45\%$     &  $81.61\%$         &  $68.57\%$     &    $90.85\%$   &   $89.13\%$    & $66.82\%$    & $86.50\%$  \\
    ResNet+OCNN \cite{wang2020orthogonal} &   $91.08\%$&  $88.50\%$     &   $88.86\%$    &  $87.08\%$     & $77.75\%$      &  $67.51\%$     &  $76.60\%$     &  $80.71\%$         &  $70.29\%$     &    $90.01\%$   &   $88.01\%$    & $65.73\%$    & $88.04\%$\\
    ResNet+hard constraints&   $90.75\%$&  $83.79\%$     &   $88.41\%$    &  $84.06\%$     & $80.80\%$      &  $69.74\%$     &  $75.91\%$     &  $85.99\%$         &  $69.84\%$     &    $91.76\%$   &   $90.88\%$    & $72.52\%$    & $88.41\%$  \\
    ResNet+WaveCNet \cite{2021ITIP...30.7074L}&  $92.18\%$&  $89.93\%$     &   $88.13\%$    &  $87.89\%$     & $75.39\%$      &  $69.23\%$     &  $80.46\%$     &  \textbf{86.23}$\boldsymbol{\%}$         &  $70.08\%$     &    $89.95\%$   &   $84.96\%$    & $66.43\%$    & $86.14\%$  \\
    \textbf{TAOTF-ResNet (Ours)} &  \textbf{92.53}$\boldsymbol{\%}$&  \textbf{93.30}$\boldsymbol{\%}$     &   \textbf{92.12}$\boldsymbol{\%}$    &  \textbf{92.66}$\boldsymbol{\%}$     & \textbf{89.76}$\boldsymbol{\%}$      &  \textbf{82.84}$\boldsymbol{\%}$     &  \textbf{84.87}$\boldsymbol{\%}$     &  $86.00\%$         &  \textbf{76.93}$\boldsymbol{\%}$     &    \textbf{92.66}$\boldsymbol{\%}$   &   \textbf{91.76}$\boldsymbol{\%}$    & \textbf{88.38}$\boldsymbol{\%}$    & \textbf{92.84}$\boldsymbol{\%}$  \\
    \cline{0-13}
    
    ViT3&  $90.99\%$&$90.10\%$     &   $89.49\%$    &  $87.23\%$     & $74.47\%$      &  $63.80\%$     &  $74.02\%$     &  $70.40\%$              &  $66.34\%$     &    $89.47\%$   &   $88.35\%$    & $73.10\%$    & $83.15\%$  \\
    ViT3+orth-initialization&   $89.92\%$&  $87.95\%$     &   $87.91\%$    &  $85.19\%$     & $75.39\%$      &  $63.06\%$     &  $71.92\%$     &  $77.00\%$         &  $63.53\%$     &    $87.59\%$   &   $88.25\%$    & $72.07\%$    & $80.43\%$  \\
    ViT3+hard constraints&   $87.62\%$&  $86.32\%$     &   $88.04\%$    &  $85.24\%$     & $75.45\%$      &  $63.10\%$     &  $72.61\%$     &  $76.30\%$         &  $69.54\%$     &    $87.23\%$   &   $88.92\%$    & $71.26\%$    & $81.52\%$  \\
    ViT3+SRIP&   $91.07\%$&  $90.12\%$     &   $90.55\%$    &  $90.19\%$     & $70.95\%$      &  $68.24\%$     &  $73.22\%$     &  $74.68\%$      
     &  $66.33\%$     &    $91.67\%$   &   $90.21\%$    & $69.56\%$    & $89.69\%$  \\
    ViT3+satge2 (self-attention layers)&   $92.60\%$&  $92.32\%$     &   $93.04\%$    & $89.52\%$&  $72.92\%$     & $65.52\%$      &  $76.93\%$     &  $79.53\%$     &  $64.49\%$         &  $92.66\%$     &    $93.57\%$   &   $61.08\%$    & $89.61\%$      \\
    ViT3+satge2 (transformer blocks)&   $95.74\%$&  $93.39\%$     &   $94.38\%$    &  $93.75\%$     & $75.45\%$      &  \textbf{75.79}$\boldsymbol{\%}$     &  $75.75\%$     &  $79.31\%$         &  $65.43\%$     &    $93.84\%$   &   $94.38\%$    & $67.30\%$    & $92.78\%$  \\
    ViT3+satge2 (patch embdding)&   $95.11\%$&  $93.84\%$     &   $93.75\%$    & $94.03\%$  &  $72.94\%$     & $69.11\%$      &  $76.99\%$     &  $80.19\%$     &  $63.98\%$         &  $94.47\%$     &    $92.84\%$   &   $61.65\%$    & $91.39\%$     \\

    \textbf{TAOTF-ViT3 (Ours)} &   \textbf{95.87}$\boldsymbol{\%}$&  \textbf{95.87}$\boldsymbol{\%}$     &   \textbf{95.81}$\boldsymbol{\%}$    &  \textbf{95.82}$\boldsymbol{\%}$     & \textbf{86.23}$\boldsymbol{\%}$      &  $75.39\%$     &  \textbf{80.92}$\boldsymbol{\%}$     &  \textbf{87.14}$\boldsymbol{\%}$         &  \textbf{74.94}$\boldsymbol{\%}$     &    \textbf{95.87}$\boldsymbol{\%}$   &   \textbf{95.56}$\boldsymbol{\%}$    & \textbf{74.70}$\boldsymbol{\%}$    & \textbf{95.29}$\boldsymbol{\%}$  \\
    \cline{0-13}
    ViT6&   $92.18\%$&  $88.89\%$     &   $87.17\%$    &  $88.40\%$     & $77.71\%$      &  $72.06\%$     &  $74.86\%$     &  $78.36\%$        &  $67.29\%$     &    $89.88\%$   &   $88.90\%$    & $75.78\%$    & $81.33\%$  \\

    ViT6+hard constraints&   $91.69\%$&  $77.66\%$     &   $75.30\%$    &  $80.65\%$     & $51.39\%$      &  $50.33\%$     &  $57.26\%$     &  $62.26\%$         &  $50.42\%$     &    $78.11\%$   &   $75.75\%$    & $52.78\%$    & $76.06\%$  \\
    ViT6+SRIP&   $93.32\%$&  $88.68\%$     &   $89.58\%$    &  $90.40\%$     & $71.89\%$      &  $66.03\%$     &  $76.39\%$     &  $74.79\%$      
     &  $66.18\%$     &    $89.04\%$   &   $88.77\%$    & $69.72\%$    & $85.18\%$  \\
 
    \textbf{TAOTF-ViT6 (Ours)} &   \textbf{94.06}$\boldsymbol{\%}$&  \textbf{94.05}$\boldsymbol{\%}$     &   \textbf{93.12}$\boldsymbol{\%}$    &  \textbf{93.51}$\boldsymbol{\%}$     & \textbf{78.82}$\boldsymbol{\%}$      &  \textbf{76.30}$\boldsymbol{\%}$     &  \textbf{76.45}$\boldsymbol{\%}$     &  \textbf{79.62}$\boldsymbol{\%}$         &  \textbf{69.82}$\boldsymbol{\%}$     &    \textbf{93.30}$\boldsymbol{\%}$   &   \textbf{92.84}$\boldsymbol{\%}$    & \textbf{76.48}$\boldsymbol{\%}$    & \textbf{90.67}$\boldsymbol{\%}$  \\
    \cline{0-13}
        \hline
        
    \multicolumn{14}{ c }{\textbf{Experiment on Skin Lesion Classification}}\\
    \hline
    \multicolumn{1}{c|}{ResNet50} &  $92.89\%$& $87.25\%$&$86.21\%$  &   $86.21\%$    & $87.17\%$& $86.38\%$ &    $86.54\%$   &  $87.25\%$            &  $59.62\%$ &$84.48\%$    &  $81.94\%$   &$81.93\%$  & $79.93\%$  \\
    ResNet50+OCNN &   $92.74\%$&  $85.27\%$     &   $86.29\%$    &  $86.39\%$     & $87.12\%$      &  $85.24\%$     &  $86.71\%$     &  $85.93\%$         &  $59.85\%$     &    $84.92\%$   &   $82.36\%$    & $82.42\%$    & $78.23\%$  \\
    ResNet50+WaveCNet &   $90.04\%$&  $88.00\%$     &   $88.06\%$    &  $87.88\%$     & $88.07\%$      &  $87.99\%$     &  $85.41\%$     &  $86.84\%$         &  $61.86\%$     &    $85.49\%$   &   $83.21\%$    & $80.07\%$    & $81.00\%$  \\
    ResNet50+SRIP &   $92.53\%$&  $86.24\%$     &   $85.45\%$    &  $84.38\%$     & $86.80\%$      &  $86.24\%$     &  $84.06\%$     &  $86.23\%$         &  $57.75\%$     &    $83.99\%$   &   $82.57\%$    & $80.69\%$    & $78.38\%$  \\

    ResNet50+hard constraints &  $91.46\%$&  $87.97\%$     &   $87.34\%$    &  $89.14\%$     & $89.17\%$      &    $88.42\%$     &  $88.62\%$     &   $88.02\%$        &    $61.69\%$   &   $86.21\%$    & $86.26\%$    & $84.75\%$& $82.27\%$      \\

    \textbf{TAOTF-ResNet50 (Ours)} &   \textbf{94.08}$\boldsymbol{\%}$    &  \textbf{94.06}$\boldsymbol{\%}$     &   \textbf{92.26}$\boldsymbol{\%}$    &  \textbf{94.02}$\boldsymbol{\%}$     & \textbf{93.69}$\boldsymbol{\%}$      &  \textbf{92.63}$\boldsymbol{\%}$     &  \textbf{91.72}$\boldsymbol{\%}$     &  \textbf{94.36}$\boldsymbol{\%}$     &  \textbf{63.54}$\boldsymbol{\%}$     &    \textbf{91.29}$\boldsymbol{\%}$   &   \textbf{90.68}$\boldsymbol{\%}$    & \textbf{88.12}$\boldsymbol{\%}$    & \textbf{86.95}$\boldsymbol{\%}$  \\
    \cline{0-13}

    \hline
    \multicolumn{14}{ c }{\textbf{Experiment on Glaucoma Detection}}\\
    \hline 
        \multicolumn{1}{c|}{ResNet} &  $92.97\%$& $88.67\%$&$90.67\%$  &   $88.50\%$    & $89.50\%$& $87.50\%$ &    $90.33\%$   &  $89.00\%$          &  $75.17\%$ &$89.67\%$    &  $90.17\%$   &$76.50\%$  & $74.00\%$  \\

    ResNet+SRIP &   $94.00\%$&  $92.50\%$     &   $90.50\%$    &  $89.00\%$     & $89.00\%$      &  $90.50\%$     &  $89.00\%$     &  $86.00\%$         &  $77.67\%$     &    $88.67\%$   &   $89.50\%$    & $82.17\%$    & $80.00\%$  \\
    ResNet+OCNN &   $93.83\%$&  $87.50\%$     &   $85.50\%$    &  $88.50\%$     & $88.50\%$      &  $88.50\%$     &  $88.50\%$     &  $87.17\%$         &  $75.67\%$     &    $85.67\%$   &   $88.80\%$    & $82.17\%$    & $74.67\%$\\
    ResNet+hard constraints&  $92.50\%$&  $91.00\%$     &   $89.00\%$    &  $90.17\%$     & $91.00\%$      &    $86.00\%$     &  $90.37\%$     &   $87.50\%$         &    $78.33\%$   &   $90.50\%$    & $90.17\%$    & $86.70\%$& $81.97\%$      \\

    \textbf{TAOTF-ResNet (Ours)} &   \textbf{94.67}$\boldsymbol{\%}$   &  \textbf{94.50}$\boldsymbol{\%}$     &   \textbf{94.00}$\boldsymbol{\%}$    &  \textbf{93.97}$\boldsymbol{\%}$     & \textbf{93.92}$\boldsymbol{\%}$      &  \textbf{93.17}$\boldsymbol{\%}$     &  \textbf{93.67}$\boldsymbol{\%}$     &  \textbf{93.77}$\boldsymbol{\%}$      &  \textbf{84.00}$\boldsymbol{\%}$     &    \textbf{93.97}$\boldsymbol{\%}$   &   \textbf{93.67}$\boldsymbol{\%}$    & \textbf{91.09}$\boldsymbol{\%}$    & \textbf{92.84}$\boldsymbol{\%}$  \\
    \cline{0-13}
    \cline{0-13}
    \end{tabular}%
    }
    \caption{Experiment on Medical Datasets. We mainly used the ViT3 model to conduct ablation experiments on the noisy Kaggle APTOS 2019, and the experiment results confirmed the role of each component of TAOTF. ``orth-initialization" means that we select orthogonal initialization weights for training. ``hard constraints" means that we impose hard orthogonal constraints (retraction-based manifold optimization algorithm) on model layers.}
  \label{tab:2019corrputed}%
    
\end{table*}%
\end{center}
\section{Experiments}\label{sec:experim_data}

In this section, to evaluate the efficiency of the proposed TAOTF framework, we conduct several experiments of the TAOTF-based DNN models on various datasets, including natural and medical ones. We implement these models on top of the deep learning framework PyTorch. Unless otherwise stated, the experimental results are measured in Top-1 Accuracy.

\subsection{Experiments on Kaggle APTOS 2019}

\subsubsection{Dataset}

We first use the public dataset \emph{Kaggle APTOS 2019}, which was collected by the Aravind Eye Hospital in India's rural areas, to evaluate the proposed TAOTF-CNN and TAOTF-ViT models.

This dataset contains 3662 retinal images, and the labels were provided by the clinicians who rated the development of Diabetic retinopathy (DR) in each image by a scale of ``0, 1, 2, 3, 4", meaning ``\emph{no DR}", ``\emph{mild}", ``\emph{moderate}", ``\emph{severe}" and ``\emph{proliferative DR}", respectively. Note that this dataset doesn't have equal distributions among the different classes.
For example, it has far more normal data with the label ``0" than other classes.
We randomly shuffle the entire dataset into three
subgroups, \emph{i.e.}, training (70$\%$), validation (10$\%$), and testing (20$\%$). 

\subsubsection{Image Preprocessing}

As different fundus images have different length-width ratios, and the width of different black edges around the eyeball is also different, we can not straightly resize the images based on their sizes.
Therefore, in this experiment, we resize the fundus images based on the eyeball radius ($224 \times 224$), and then use the feature enhancement method. 
In this process, the difference between the original image and the Gaussian blurred one (equivalent to the background) is used to enhance the feature.

\subsubsection{Models and Settings}
We choose the ResNet18, ViT3 (3 transformer blocks), and ViT6 (6 transformer blocks) models to test the robustness performances of TAOTF-based models on the Diabetic Retinopathy classification task. For training these models, the total epoch of the training is 200. We start the learning rate with $lr=3 \times 10^{-5}$, with weight decay 1e-4. The weight $\lambda$ of the regularization loss is $10^{-3}$, the model is trained using Adam, and the batch size is 8. In the above training process, we use CrossEntropyLoss (CE) for the criterion.

\subsubsection{Experimental Results on Clean Dataset} 
Proper orthogonality constraints can help fully utilize
the model capacity. We also have conducted ablation experiments for each part of the framework based on the ViT3 model. 

The experimental results show that adding soft orthogonal constraints to all layers of DNN can help improve the model performance. The experimental results on the clean dataset are summarized in \cref{2019kaggle}. It can be seen that the proposed TAOTF-based models have better performances than other methods.

\subsubsection{Noisy Dataset for Testing Roubustness}To evaluate the robustness of TAOTF-based models compared with other methods, we ask for ophthalmologists and conclude 13 common data corruption of fundus images and classified into 4 types. Then we simulated these possible data challenges to build a noisy test set for test model robustness performances. For example, the geometric transformation could test model performance to the position deviation, viewing angle deviation, and data size deviation, the spatial transformation could test the model performance to the change of light, color, contrast, and brightness, and finally test the model performance to fuzzy images and images with more noise.
\subsubsection{Experimental Results on Noisy Data}

We also have conducted ablation experiments for each part of the framework based on the ViT3 model. The experimental results show that all parts of our framework have improved the robustness of the model to a certain extent. And the results of the test set are summarized in \cref{tab:2019corrputed}. It can be seen that the proposed TAOTF-based models have better robustness performances than other existing methods in this task. Because proper orthogonality makes each DNN layer more approximately a 1-Lipschitz function, yields representations that are robust and less sensitive to perturbations.
\begin{table}[!t]
\newcommand{\tabincell}[2]{\begin{tabular}{@{}#1@{}}#2\end{tabular}}
\centering

\resizebox{\linewidth}{!}{
\begin{tabular}{cccc}
\toprule

     & \tabincell{c}{\textbf{Accuracy}} & \tabincell{c}{\textbf{Precision}} &  \tabincell{c}{\textbf{Recall}}
     \\
\midrule
\midrule
 \tabincell{c}ResNet & $91.17\%$ & $91.06\%$  & $90.93\%$ \\
 \midrule
 \tabincell{c}ResNet+hard constraints   & $90.75\%$ & $90.62\%$  & $90.53\%$  \\
  \midrule
 \tabincell{c}ResNet+SRIP    & $91.17\%$ & $91.10\%$  & $91.17\%$  \\
 \midrule
 \tabincell{c}ResNet+OCNN    & $91.08\%$ & $91.26\%$  & $91.08\%$  \\
 \midrule
 \tabincell{c}ResNet+WaveCNet    & $92.18\%$ & $92.34\%$  & $92.18\%$  \\
 \midrule
\tabincell{c}{\textbf{TAOTF-ResNet (Ours)}}    & \textbf{92.53}$\boldsymbol{\%}$ & \textbf{92.46}$\boldsymbol{\%}$ & \textbf{92.53}$\boldsymbol{\%}$  \\
\bottomrule
\midrule
\tabincell{c}ViT3   & $90.99\%$ & $90.99\%$  & $90.83\%$  \\
\midrule
\tabincell{c}ViT3+orth-initialization   & $89.92\%$ & $89.85\%$  & $89.49\%$  \\
\midrule
\tabincell{c}ViT3+hard constraints   & $87.62\%$ & $87.58\%$  & $87.29\%$  \\
\midrule

\tabincell{c}ViT3+SRIP   & $91.07\%$ & $91.09\%$  & $91.02\%$  \\
\midrule
\tabincell{c}ViT3+stage2 (only self-attention layers)   & $92.60\%$ & $92.78\%$  & $92.21\%$  \\
\midrule
\tabincell{c}ViT3+stage2 (only transformer blocks)   & $95.74\%$ & $95.78\%$  & $95.71\%$  \\
\midrule
\tabincell{c}ViT3+stage2 (only patchembedding)   & $95.11\%$ & $95.16\%$  & $95.00\%$  \\
\midrule
\tabincell{c}{\textbf{TAOTF-ViT3 (Ours)}}    & \textbf{95.87}$\boldsymbol{\%}$ & \textbf{95.81}$\boldsymbol{\%}$ & \textbf{95.80}$\boldsymbol{\%}$  \\
\bottomrule
\end{tabular}
}
\caption{Experiment on clean Kaggle APTOS 2019 (Ablation experiments). Our framework impose models with proper orthogonality constraints, which can improve task performance. }
\label{2019kaggle}
\end{table}

\begin{center}
\begin{table*}[!htbp]
  \centering

    \resizebox{\linewidth}{!}{\begin{tabular}{c|c|ccc|ccc|ccc|ccc}
        \hline
    \multicolumn{14}{ c }{\textbf{Experiment on CIFAR-100}}\\
    \hline

              & Clean      &       &    Noise   &       &     & Blur       &       &       &  Weather     &       &       & Digital \\
         \cline{0-13}
          \multicolumn{1}{c|}{Methods}& \multicolumn{1}{c|}{Clean}& \multicolumn{1}{c}{Gaussian.} & \multicolumn{1}{c}{ISO.} & \multicolumn{1}{c|}{Multiplicative.} & \multicolumn{1}{c}{Gauss.} & \multicolumn{1}{c}{Median}
          & \multicolumn{1}{c|}{Motion} & \multicolumn{1}{c}{Optical}  & \multicolumn{1}{c}{Rotate} &\multicolumn{1}{c|}{RGB} & \multicolumn{1}{c}{Bright} & \multicolumn{1}{c}{Frog}& \multicolumn{1}{c}{Saturation} \\
          \cline{0-13}
    \multicolumn{1}{c|}{WideResnet} &$68.87\%$     & $35.87\%$&$21.75\%$  &   $27.77\%$    & $03.86\%$& $13.25\%$ &    $19.25\%$   &  $26.50\%$     &   $23.34\%$ &$51.93\%$    &  $52.19\%$   &$54.07\%$  & $47.59\%$  \\
    WideResnet+SRIP &   $70.97\%$    &  $45.06\%$     &   $33.02\%$    &  $40.76\%$     & $07.40\%$      &  $19.29\%$     &  $29.31\%$     &  $37.53\%$     &  $41.24\%$     &    $59.06\%$   &   $58.92\%$    & $60.95\%$    & $55.19\%$  \\

   WideResnet+hard constraints&  $71.04\%$&  $53.40\%$     &   $38.24\%$    &  $46.72\%$     & $17.94\%$      &    $27.70\%$     &  $41.60\%$     &   $52.18\%$         &    $42.85\%$   &   $64.54\%$    & $64.06\%$    & $65.91\%$& $60.20\%$      \\
    \textbf{TAOTF-WideResnet (Ours)} &\textbf{71.09}$\boldsymbol{\%}$&  \textbf{61.06}$\boldsymbol{\%}$     &  
    \textbf{45.66}$\boldsymbol{\%}$    &  \textbf{56.02}$\boldsymbol{\%}$     & \textbf{20.24}$\boldsymbol{\%}$      &  \textbf{33.89}$\boldsymbol{\%}$     &  \textbf{47.15}$\boldsymbol{\%}$     &  \textbf{57.56}$\boldsymbol{\%}$        &  \textbf{59.04}$\boldsymbol{\%}$     &    \textbf{69.52}$\boldsymbol{\%}$   &   \textbf{69.21}$\boldsymbol{\%}$    & \textbf{71.01}$\boldsymbol{\%}$    & \textbf{65.35}$\boldsymbol{\%}$    \\
    \cline{0-13}
    \cline{0-13}
    \hline
    \multicolumn{14}{ c }{\textbf{Experiment on CIFAR-10}}\\
    \hline

          \cline{0-13}
    \multicolumn{1}{c|}{MobileViT} &  $83.53\%$ & $80.03\%$&$72.30\%$  &   $74.75\%$    & $69.21\%$& $69.37\%$ &    $74.73\%$   &  $80.59\%$          &  $69.15\%$ &$81.50\%$    &  $81.90\%$   &$83.06\%$  & $81.18\%$  \\

    MobileViT+SRIP  &   \textbf{84.35}$\boldsymbol{\%}$&  $79.61\%$     &   $71.99\%$    &  $74.29\%$     & $69.24\%$      &  $69.30\%$     &  $75.14\%$     &  $80.37\%$        &  $69.72\%$     &    $81.64\%$   &   $81.89\%$    & $83.27\%$    & $81.26\%$  \\

   MobileViT+hard constraints&  $77.80\%$ &  $77.19\%$     &   $69.88\%$    &  $71.64\%$     & $71.44\%$      &    $70.44\%$     &  $76.42\%$     &   $80.83\%$        &    $70.95\%$   &   $81.28\%$    & $82.51\%$    & $83.28\%$& $81.00\%$      \\
    \textbf{TAOTF-MobileViT (Ours)} &  $84.10\%$  &\textbf{81.44}$\boldsymbol{\%}$     &  
    \textbf{75.33}$\boldsymbol{\%}$    &  \textbf{77.94}$\boldsymbol{\%}$     & \textbf{75.10}$\boldsymbol{\%}$      &  \textbf{74.32}$\boldsymbol{\%}$     &  \textbf{77.72}$\boldsymbol{\%}$     &  \textbf{82.20}$\boldsymbol{\%}$         &  \textbf{72.35}$\boldsymbol{\%}$     &    \textbf{83.49}$\boldsymbol{\%}$   &   \textbf{84.97}$\boldsymbol{\%}$    & \textbf{84.57}$\boldsymbol{\%}$    & \textbf{83.17}$\boldsymbol{\%}$    \\
    \cline{0-13}
    \cline{0-13}
 
    \end{tabular}%
    }
    
\caption{Experiment on CIFAR-100/CIFAR-10}
  \label{tab:Cifar10}%

\end{table*}%
\end{center}

\begin{figure}[!tbp]
\centering
\subfloat{\includegraphics[width=0.35\textwidth]{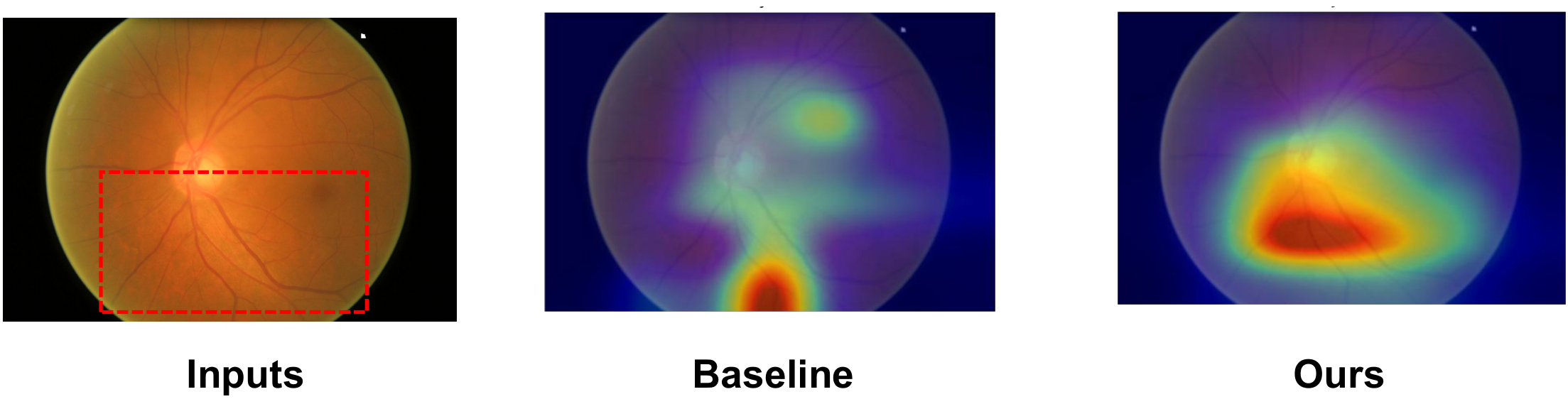}}
\caption{Glaucoma Detection Classification visualization by class activation maps (CAM) \cite{zhou2016learning}. Our framework can help models find the accurate location of lesions (key features).} 
\label{CAM}
\end{figure}
\begin{figure}[!tbp]
\centering
\subfloat{\includegraphics[width=0.4\textwidth]{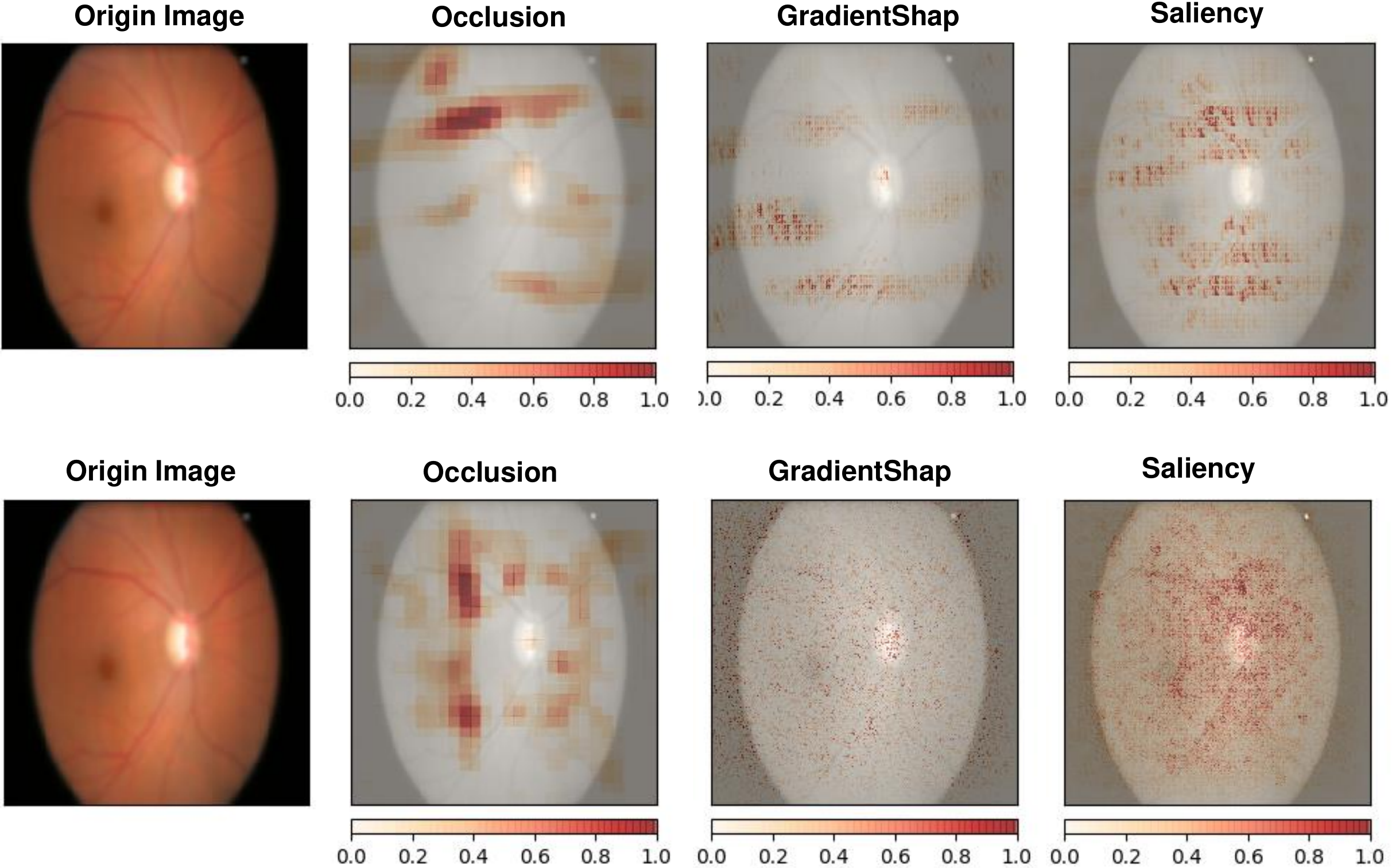}}
\caption{Glaucoma Detection Classification visualization (blur image) by Towhee, including Occlusion, GradientShap, Saliency. Our framework can help models find the accurate location of lesions (key features).} 
\label{CAM2}
\end{figure}
\subsection{Experiments on Glaucoma Detection Dataset}
\subsubsection{Experimental Setup}

For datasets, we choose a Kaggle Glaucoma Detection Dataset to test our framework performance. The dataset contains 650 images/OCT scans of the eyes. The labels were provided by clinicians who rated the Glaucoma in each image on a scale of ``0, 1", which means ``\emph{No Glaucoma}", and ``\emph{Glaucoma}" respectively. 
We randomly shuffle the entire dataset into three subgroups, \emph{i.e.}, training (70$\%$), validation (10$\%$), and testing (20$\%$). 

For the training process, in the experiments,  we choose the ResNet18 model to classify Glaucoma and test the performance of our framework. We use the Ranger with $lr=3 \times 10^{-5}$, with weight decay 1e-3, and train it for 130 epochs. The weight $\lambda$ of the regularization loss is $10^{-3}$. In the above training process, we use CE Loss as the criterion. 

\subsubsection{Experimental Results}

We compare our TAOTF with prior works. And our framework  significantly outperforms existing methods, which shows that TAOTF-based models have stronger robustness in this medical dataset. See \cref{tab:2019corrputed} for a detailed comparison. After that, our model visualization can be seen \cref{CAM} and \cref{CAM2}. 

\subsection{Experiments on Skin Lesion Classification}
\subsubsection{Experiment Setup}
For datasets, this dataset contains the training data for the ISIC 2019 challenge \cite{2019arXiv190802288C}, and datasets from previous years (2018 and 2017). \cite{2018NatSD...580161T} \cite{2016arXiv160501397G}. The dataset contains 25331 images available to classify dermoscopic images among nine diagnostic categories. The labels were provided by clinicians who rated the classification of a skin lesion in each image on a scale of ``0, 1, 2, 3, 4, 5, 6, 7, 8", which means ``\emph{Melanoma}", ``\emph{Melanocytic nevus}", ``\emph{Basal cell carcinoma}", ``\emph{Actinic keratosis}", ``\emph{Benign keratosis}", ``\emph{Dermatofibroma}", ``\emph{Vascular lesion}", ``\emph{Squamous cell carcinoma}", ``\emph{None of the above}". We randomly shuffled the entire dataset into three subgroups, \emph{i.e.}, training (80$\%$), validation (10$\%$), and testing (10$\%$).
\begin{table}[!tbp]

    \newcommand{\tabincell}[2]{\begin{tabular}{@{}#1@{}}#2\end{tabular}}
    \small
        \centering

    \begin{tabular}{cccc}
    
    \toprule
    
         & \tabincell{c}{\textbf{BCE (↓)}} & \tabincell{c}{\textbf{ACC (↑)}} &  \tabincell{c}{\textbf{F1-Score (↑)}}
         \\
    \midrule
    \midrule
     \tabincell{c}{UNet} & 0.00713 &  0.996 & 0.837  \\
     \midrule
      \tabincell{c}{UNet+SRIP} & 0.00949 &  0.996 & 0.811  \\
     \midrule    
     \tabincell{c}{\textbf{TAOTF-UNet (Ours)}}  &\textbf{0.00697}  & \textbf{0.997} &  \textbf{0.843} \\

    \bottomrule
    \end{tabular}
    \caption{ Experiment on Brain MRI segmentation. The proposed TAOTF can help improve segmentation performance.}
    \label{segmentation}
\end{table}

For training, we choose ResNet-50 to test our framework. We use the Adam with $lr=0.001$, and train it for 100 epochs. The weight $\lambda$ of the regularization loss is $10^{-4}$. In the above trainings, we use the CE Loss as the criterion. 

\subsubsection{Experimental Results}
We compare our framework TAOTF with other methods. See \cref{tab:2019corrputed} for a detailed comparison. The above results confirm that imposing proper orthogonality constraints for models has stronger robustness performances on this noisy test dataset.

\subsection{Experiments on Brain MRI segmentation}

\subsubsection{Experiment Setup}
The dataset contains brain MR images together with manual FLAIR abnormality segmentation masks. The dataset containing 3929 images was obtained from The Cancer Imaging Archive (TCIA). They correspond to 110 patients included in The Cancer Genome Atlas (TCGA) lower-grade glioma collection with at least fluid-attenuated inversion recovery (FLAIR) sequence and genomic cluster data available. We randomly shuffle the entire dataset into three subgroups, \emph{i.e.}, training (70$\%$), validation (10$\%$), and testing (20$\%$). We choose UNet \cite{ronneberger2015u}, which is composed of 10 convolution layers. Training takes 30 epochs with the TAOTF regularizer applied to the model. The weight $\lambda$ of the regularization loss is $10^{-3}$ and all of other settings retain as
standard default.

\subsubsection{Experiment Results}
We compare TAOTF-based UNet with other methods.
See \cref{segmentation} for a detailed comparison. The above results
confirm that TAOTF-based models can help improve segmentation tasks.

\begin{table}[tbp]

    \newcommand{\tabincell}[2]{\begin{tabular}{@{}#1@{}}#2\end{tabular}}
    \small
        \centering

    \begin{tabular}{cccc}
    
    \toprule
    
         & \tabincell{c}{\textbf{PSNR (↑)}} & \tabincell{c}{\textbf{SSIM (↑)}} &  \tabincell{c}{\textbf{LPIPS (↓)}}
         \\
    \midrule
    \midrule
     \tabincell{c}{CNN} & 26.697 &  0.893 & 0.092  \\
     \midrule
     
     \tabincell{c}{\textbf{TAOTF-CNN (Ours)}}  &\textbf{29.117}  & \textbf{0.942} &  \textbf{0.084} \\

    \bottomrule
    \end{tabular}
    \caption{ Experiment on Denoising CIFAR-10. Proper orthogonality constraints can help the model ignore small input perturbations and improve model performance facing low-level
tasks (\emph{e.g.}, noise, blur).}
    \label{denosing}
\end{table}

\subsection{Experiments on CIFAR-10/CIFAR-100} 
\subsubsection{Experiment Setup}
For datasets, the CIFAR-10 dataset has a total of 60000 color images. These images are $32\times32\times3$ and are divided into 10 categories, with 6000 images in each category. Among them, 50000 images are used for training, another 10000 images are used for testing. The CIFAR-100 dataset has 100 categories, with 500 training images and 100 testing images per category. 

For training, in the first dataset CIFAR-10, we choose the MobileViT-s \cite{2021arXiv211002178M}, which combined with the transformers and CNNs, to test our framework performance on the CIFAR-10 dataset. We use the AdamW with $lr=0.001$ to train it for 60 epochs. In the second dataset, we choose 28-depth WideResNet \cite{zagoruyko8wide} to test our framework performance on the CIFAR-100 dataset further. We use the Adam with $lr=0.0005$ to train it for 80 epochs. The weight $\lambda$ of the regularization loss is $10^{-4}$. In the above trainings, the criterion chosen is CE Loss with 0.1 label smoothing to reduce overfitting.
\subsubsection{Experimental Results}
We compare our framework TAOTF with other methods. See \cref{tab:Cifar10} for a detailed comparison. The above results confirm that TAOTF-based models can also resist the corruption of natural images to a certain extent.

To better understand how our framework works, we design a simple auxiliary denoising experiment on CIFAR-10. We add the same intensity of noise on CIFAR-10 and use a simple CNN model (16 layers) to denoising it. See \cref{denosing} for a detailed comparison. Our TAOTF-based models can ignore small input perturbations(\emph{e.g.}, noise, blur), enjoying robustness under noisy data.

\section{Conclusion}

According to the practical difficulties encountered in data quality, we proposed a new two-stage training framework TAOTF, which can find a trade-off between the orthogonality constraint and the main task solution set, and propose an orthogonal initialization algorithm PDOI in the first stage that can find a good starting point for orthogonal optimization. Our framework was tested both on transformers and CNNs, and the experimental results show that our framework can significantly improve the robustness performances of these models facing noisy data.

\bibliographystyle{IEEEbib}
\bibliography{strings,refs}

\end{document}